\documentclass[conference]{IEEEtran}
\IEEEoverridecommandlockouts
\usepackage{cite}
\usepackage{amsmath,amssymb,amsfonts}
\usepackage{algorithmic}
\usepackage{graphicx}
\usepackage{textcomp}
\usepackage{xcolor}
\def\BibTeX{{\rm B\kern-.05em{\sc i\kern-.025em b}\kern-.08em
    T\kern-.1667em\lower.7ex\hbox{E}\kern-.125emX}}

\usepackage[T1]{fontenc}
\usepackage[utf8]{inputenc}
\usepackage{cite}
\usepackage{bm}
\usepackage{algorithm}
\usepackage{array}
\usepackage{booktabs}
\usepackage{indentfirst}
\usepackage{amsmath}
\usepackage{amsfonts}
\usepackage{multirow}
\usepackage{booktabs}
\usepackage{ulem}
\usepackage{cite}

\usepackage{url}
\usepackage{tabu}
\usepackage{threeparttable}
\usepackage{adjustbox}
\usepackage{subfigure}
\usepackage{multirow}
\usepackage{epstopdf}

\begin{document}

\title{Graph Force Learning\\
}

	\author{
		\IEEEauthorblockN{Ke Sun$^1$, Jiaying Liu$^1$, Shuo Yu$^1$, Bo Xu$^1$, and Feng Xia$^2$}
		\IEEEauthorblockA{$^1$School of Software, Dalian University of Technology, Dalian 116620, China\\
			$^2$School of Engineering, IT and Physical Sciences, Federation University Australia, VIC 3353, Australia\\
			\{kern.sun, jiaying\_liu, y\_shuo\}@outlook.com, boxu@dlut.edu.cn, f.xia@ieee.org
		}
	}

\maketitle

\begin{abstract}
Features representation leverages the great power in network analysis tasks. However, most features are discrete which poses tremendous challenges to effective use. Recently, increasing attention has been paid on network feature learning, which could map discrete features to continued space. Unfortunately, current studies fail to fully preserve the structural information in the feature space due to random negative sampling strategy during training. To tackle this problem, we study the problem of feature learning and novelty propose a force-based graph learning model named GForce inspired by the spring-electrical model. GForce assumes that nodes are in attractive forces and repulsive forces, thus leading to the same representation with the original structural information in feature learning. Comprehensive experiments on benchmark datasets demonstrate the effectiveness of the proposed framework. Furthermore, GForce opens up opportunities to use physics models to model node interaction for graph learning.
\end{abstract}


\section{Introduction}
A network is composed of a set of nodes and edges, which is commonly employed to represent data in real world. The effective representation of network features can significantly improve the performance of downstream tasks~\cite{bengio2013representation}. For example, in knowledge graph research, effective entity feature representation can promote the establishment of more accurate relationship between entities~\cite{shang2019end}; in online social recommendation systems, capable user feature representation is conducive to accurate recommendation~\cite{wu2019neural}. Therefore, it is necessary to develop algorithms that could automatically extract features representation from the original graph, thus freeing people from tedious feature extraction works.

Network feature learning~\cite{cui2018survey} is critical for network analysis tasks, such as node clustering~\cite{tang2015line}, label prediction~\cite{grover2016node2vec}, and link prediction~\cite{wang2020Motif}. It could map desired information of vertices or edges to low dense feature vectors in a continuous feature space. The features extracted from the network simultaneously preserve network structural information, node attributes information, and promote users to deeply understand hidden characteristics of networked data. What calls for special attention is that most existing network feature learning algorithms could not fully preserve original features of network structure due to the random sampling of negative nodes~\cite{grover2016node2vec}. Specifically, unconnected nodes have similar representation in the feature space due to the mechanism of most current methods. For example, Node2vec~\cite{grover2016node2vec} and DeepWalk~\cite{perozzi2014deepwalk} algorithms are based on skip-gram model~\cite{morin2005hierarchical,mikolov2013efficient,mikolov2013distributed}. They rely on negative sampling to calculate nodes similarity. Therefore, negative samples that are not selected might be very close to the target nodes in the feature space. Matrix factorization based methods, such as GraRep~\cite{cao2015grarep}, HOPE~\cite{ou2016asymmetric}, cannot get rid of the disadvantages of matrix decomposition, e.g., computational complexity. Recent advancements in network feature learning mainly focus on deep learning, such as graph convolutional networks~\cite{ma2019graph}, graph attention networks~\cite{kosaraju2019social}, and graph capsule networks~\cite{hahn2019self}. In these studies, the total loss function and stochastic gradient descent (SGD) are employed to optimize weight matrices for minimizing the loss function. A large amount of parameter matrices of perceptrons are required to be updated, but still cannot resolve the problem of gradient explosion.

\begin{figure}[t]
  \centering
  \includegraphics[width=3.5 in,height=1.2 in]{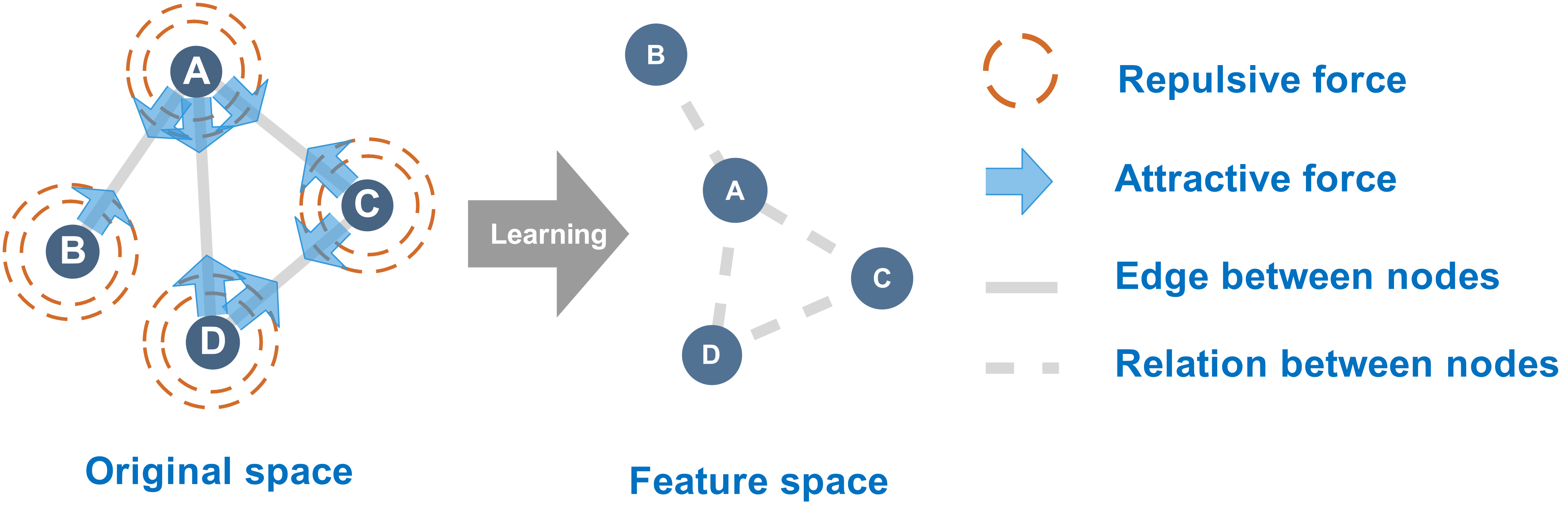}
  \caption{The principle of GForce. The dash circle represents the repulsive force between nodes. Two nodes will move far away with the force. The vector arrow represents an attractive force between two nodes. Finally, nodes will learn the right location (vector) in the feature space.}
  \label{electronic_spring}
\end{figure}

In this work, we propose a non-neural network feature learning algorithm, namely GForce, that can fully preserve the structure in the feature space. GForce is inspired by the spring-electrical model~\cite{ko2020flow}, which simulates the motion of particles under force. Such idea has been successfully applied to graph visualization~\cite{arleo2018distributed,haleem2019evaluating,jenny2017force} in 2D and 3D space, aiming at presenting the original network topology of graph. The similar and connected nodes tend to be located in the similar position or belonging to the same cluster. Therefore, we believe that the positions of nodes in 2D or 3D space can be regarded as vectors of nodes in feature space, so similar embedding vectors can represent similar information of nodes. We adopt the idea of spring-electrical model, and extend embedded dimensions to any dimensions. Our goal is to fully extract structural features from the observed network rather than using any hand-engineered features. We assume that each node is repulsive to other nodes but attractive to neighbors. The two connected nodes will be relatively close to each other, otherwise, nodes will be far from each other in feature space. The core idea of GForce can be simply illustrated as Fig~\ref{electronic_spring}. In addition, GForce can directly learn the vectors of nodes without using SGD. Every node calculates its own vector, and they move through attractive relations and repulsive relations in the feature space. Since every node computes its own vector, the algorithm could run in a parallel model with multi-threads and multi-processes.

We evaluate the performance of GForce against several baselines on five real-world datasets (i.e., Cora\footnote{http://linqs.cs.umd.edu/projects//projects/lbc},
Citeseer\footnote{https://csxstatic.ist.psu.edu/downloads/data}, Wikipedia\footnote{http://www.cs.umd.edu/~sen/lbc-proj/LBC.html}, WebKB\footnote{http://www-2.cs.cmu.edu/~webkb/}, 20newsgroups\footnote{http://qwone.com/~jason/20Newsgroups/}). Experimental results show that our algorithm achieves better performance in label prediction and graph visualization in comparison with the state-of-the-art baselines (i.e., DeepWalk~\cite{perozzi2014deepwalk}, LINE~\cite{tang2015line}, Node2vec~\cite{grover2016node2vec}). Our major contributions can be summarized as follows:
\begin{itemize}
\item We propose a novel graph force learning algorithm based on the spring-electrical model. The proposed GForce can sufficiently preserve structural information, which fills the gap of network feature learning.
\item GForce could run in parallel with threads or multi-processes, that make full use of computing resources. The representation process of GForce depends on network structure information. Therefore, GForce owns the advantage of expandability in graph structure based application scenarios.
\item We evaluate the GForce on real-world datasets in the tasks of label prediction and graph visualization. Experimental results show the effectiveness and efficiency of our proposed algorithm.
\end{itemize}

The rest of this paper is organized as follows. We first discuss related works in Section~\ref{sec:related_works}. Then, we present related notations and definitions of graph feature learning in Section~\ref{sec:preoblem_def}. And then, we introduce the GForce in Section~\ref{sec:netrelation}. In the following, we outline the experiments and present the results in Section~\ref{sec:experiments}. Finally, we conduct our works in Section~\ref{sec:conclusion}.

\section{Related Work}\label{sec:related_works}
Network feature learning is always an important research problem in network science. Amount of previous studies have proved its validity for applications, such as link prediction, label prediction, and network visualization. There have been a large amount of network embedding models proposed in recent years, such as linear methods PCA~\cite{lever2017points}, and non-linear methods including IsoMap~\cite{tenenbaum2000global}, multidimensional scaling (MDS)~\cite{saeed2018survey}, and LLE~\cite{cao2016supervised}. These methods are mainly applied to network embedding and dimensionality reduction. However, these methods have disadvantages in both computational and statistical performance.

In recent years, some representative studies have been devoted to other aspects of this field. For example, GraRep~\cite{cao2015grarep}, HOPE~\cite{ou2016asymmetric}, TADW~\cite{yang2015network} are matrix factorization methods. They utilize matrix factorization to obtain nodes embeddings. DeepWalk~\cite{perozzi2014deepwalk}, Node2vec~\cite{grover2016node2vec} are established based on the Skip-gram model~\cite{mikolov2013efficient,morin2005hierarchical,mikolov2013distributed}, which is a kind of natural language processing model. DeepWalk utilizes a truncated random walk to capture the network structure information. Node2vec improves the sampling strategy with walking preference parameters to maintain the local or global characteristics of the graph. Besides, some work focuses on edge modeling, such as LINE~\cite{tang2015line} and TLINE~\cite{zhang2016tline}.

Taking advantage of the ability to extract hidden features in deep learning, recent advancements in network feature learning are mainly focused on deep learning based models. These deep learning technologies such as convolutional neural networks~\cite{choy20194d}, autoencoder~\cite{yang2019deep}, attention neural networks~\cite{kong2019weakly}, capsule neural networks~\cite{vijayakumar2019comparative}, and reinforcement learning~\cite{cobbe2019quantifying} are leveraged to network embedding~\cite{zhang2020deep}, such as VGAE~\cite{shi2020effective}, Shifu2~\cite{liu2019shifu2}, graph convolutional networks~\cite{yao2019graph,xu2020multivariate}, graph attention networks~\cite{song2019session,kosaraju2019social}, graph capsule networks~\cite{ma2019disentangled,ahmed2019star}, graph generative network~\cite{grover2019graphite} and graph reinforcement learning~\cite{wang2019adaptive}. These deep learning based models rely on a total loss function and utilize SGD to optimize weight matrices to minimize the loss function. Hence, these methods need to update a large number of weight matrices by using SGD.

Different from all these graph feature learning algorithms, we abandon the way of constructing complex graph neural-based model and the mechanism of message propagating between nodes, which are often utilized by recent advancements in network feature learning. Most of them rely on SGD to learn the feature vectors. We novelly involve physical model to graph feature learning, and propose the GForce. GForce enables each node to have the learning ability to calculate similarity or heterogeneity to others. This is why GForce can be computed in parallel without depending on a single loss function and SGD. Nodes can independently learn the feature vector through direct and concise operations according to the structure information while it could fully preserve original structure information in feature space.

\section{Preliminaries}\label{sec:preoblem_def}
We focus on network (graph) structure data. In this section, we formally give the related definitions of graph, network embedding, label prediction, etc.
\\

DEFINITION 1: (Graph). A graph is defined as a set of nodes and edges, $G = (V,E,Y)$, where $V=\{v_1,v_2,\ldots,v_n\}$ is a set of nodes. Each node is usually associated with labels and attributes $Y$. The notation $E \subseteq (V \times V)$ is a set of edges between nodes. Each edge $e \in E$ connects a pair of nodes $e = (i,j)$. Edges are associated with weight $w_{ij} > 0$, which represents the strength of relations between two nodes. For undirected graph $G$, $e_{ij}=e_{ji}$. Otherwise, $e_{ij}\neq e_{ji}$.

In practice, graph is a kind of complex structure data. A graph can be directed (e.g., citation networks), undirected (e.g., social network), homogeneous (e.g., protein network), and heterogeneous (e.g., webpage network). Weights between nodes can be real or binary. This paper mainly studies isomorphic graphs.
\\

DEFINITION 2: (K-order Proximity). The K-order proximity is used to weigh the relation between two nodes in network, where $K \subseteq N+$ is a set of positive integers. For instance, K=1 or K=2 is corresponding to the first-order proximity and second-order proximity of nodes. The first-order (K=1) proximity is evaluated by the weight of the edge connected by two nodes. The second-order (K=2) proximity is determined by directed neighborhoods of two compared nodes. Specifically, given neighbors $N_i = \{s_i1,\ldots,s_i|V|\}$ of node $i$ and $N_j = \{s_j1,\ldots,s_j|V|\}$ of node $j$, the second-order proximity between $i$ and $j$ is calculated as the similarity between $N_i$ and $N_j$.

K-order proximity is based on the assumption that nodes tend to have similar information and attributes if they have a direct or indirect connection. For example, people who are friends with each other are likely to have similar interests in social networks. Two persons who have similar friends may also have the same preferences in shopping or other habits.
\\

DEFINITION 3: (Network Embedding). Given a network $G$, the aim of network embedding is to learn a function that can map nodes to a continuous feature space $f: V \mapsto \mathbf{R}^d$, where $d \ll |V|$. In the feature space, the distance between nodes approximates to the real distribution of the original space, maintaining first-order proximity, second-order proximity or high-order proximity.
\\

DEFINITION 4: (Label Prediction in Feature Space). Given feature vectors $\mathbf{U}$ generated from feature mapping function $f$ by graph $G = (V,E,Y)$, we aim to learn a hypothesis $H$, that could aggregate labels to unseen data $H: \mathbf{U} \mapsto \mathbf{Y}$.
\\

In the following, we introduce a force-based graph learning model that could fully preserves structure information.

\begin{figure}[t]
  \centering
  \includegraphics[width=3.4in,height=2.3 in]{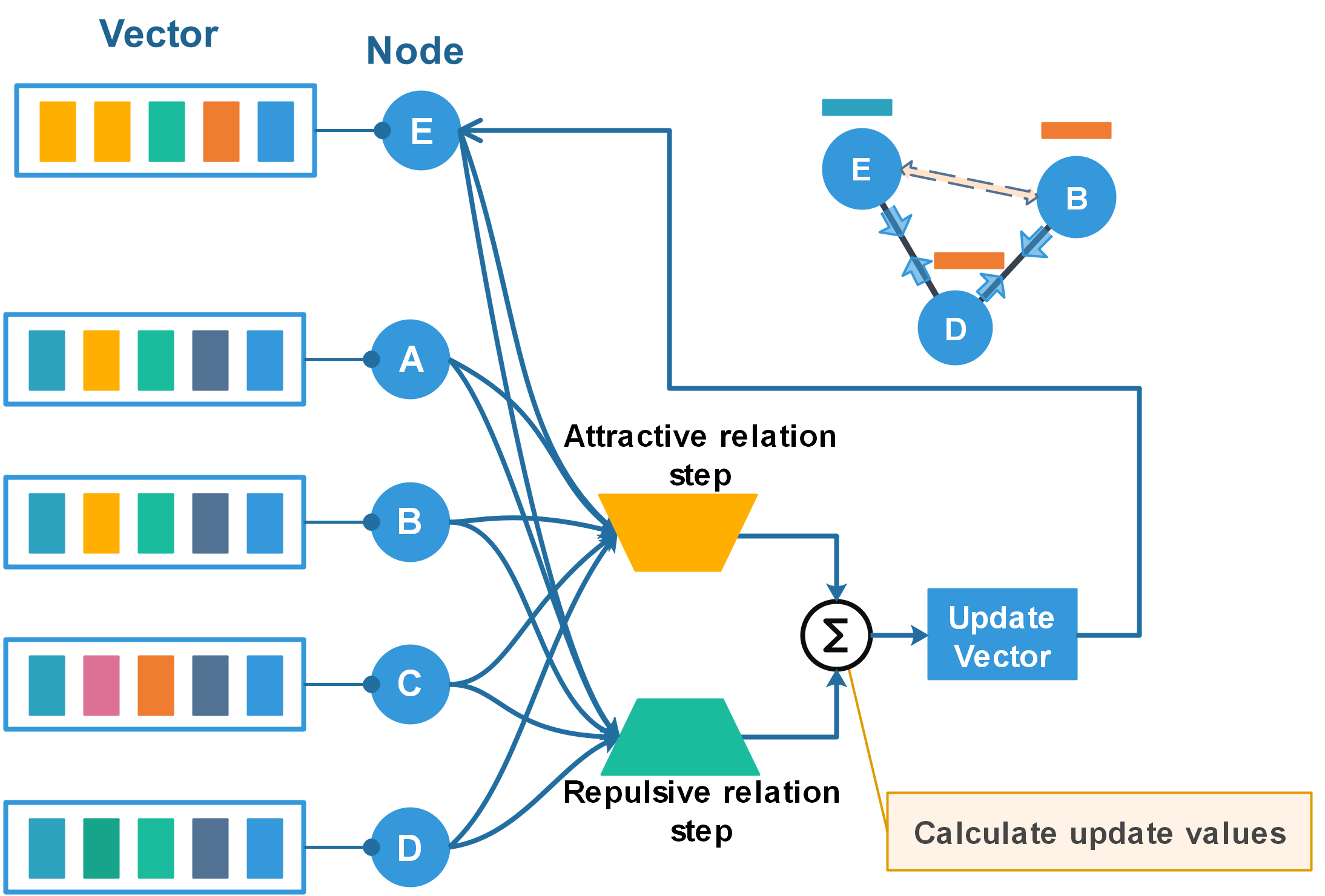}
  \caption{The framework of GForce.}
  \label{NetForce_framework6}
\end{figure}
\section{Graph Force Learning}\label{sec:netrelation}
The algorithm contains two main steps: attractive relation step and repulsive relation step similar to spring-electrical model that has attractive and repulsive components. Attractive relation step could make similar nodes move to closer positions. Repulsive relation component makes different nodes stay away from each other. The input of GForce is an undirected (un)weighted graph. Given a graph $G = \{V,E\}$, the extracted feature vectors of $G$ is $\mathbf{U} \in \mathbf{R}^{|V|\times n}$, where the dimension of feature space: $n \gg 3$. The update process of feature vectors is expressed as the following equation:
\begin{equation}
\mathbf{U} = \mathbf{U} + h*(\nabla F_a+ \nabla F_r),
\end{equation}
where $\nabla F_a$ is the update of attractive relation step, $\nabla F_r$ is the update of repulsive relation step. The notation $h$ is a parameter to control the learning speed. It can be a constant or a self-adjusting variable. In this paper, we adopt the calculation method proposed by Weinred et al.~\cite{weinreb2018spring}, which could adapt to both global and local learning speed on the graph. The feature vectors $\mathbf{U}$ update in each iteration. We define the distance between nodes $i$ and $j$ is $\mathbf{d}_{ij} = \mathbf{u}_i-\mathbf{u}_j$. The update of attractive relation of node $i$ is expressed as following:
\begin{equation}
\nabla f_a(i) = -p*\hat{w}_{ij}*(\mathbf{u}_i-\mathbf{u}_j),
\label{con:f_a}
\end{equation}
where $\mathbf{u}_i$ is an feature vector of nodes $i$. $p$ is the learning rate, which controls the learning speed. It can be set to a constant (such as $p=1$). $\hat{w}_{ij} \in \hat{\mathbf{W}}$ is a positive link weight between nodes $i$ and $j$ from a dataset. The repulsive relation strength is inversely proportional to Euler distance $\mathbf{d}$. However, when $\mathbf{d}$ approaches zero ($\mathbf{d} \to \mathbf{0}$), $\nabla f_r$ approaches infinite ($\nabla f_r \to \infty$). To overcome this weakness, we design a bias repulsive relation method. The update of it is expressed as:
\begin{equation}
\nabla f_r(i) = q*\tilde{w}_{ij}*\frac{(\mathbf{u}_i-\mathbf{u}_j)}{||\mathbf{u}_i-\mathbf{u}_j|+\mathbf{b}|^2},
\label{con:f_r}
\end{equation}
where $\mathbf{b}$ is a distance bias, which can be set to a small positive value, for example, $\mathbf{b} = \{0.01\}_1^n$. $q$ has similar function to $p$, which controls the learning speed. $\tilde{w}_{ij}$ is a negative weight between two nodes divergence. We can regard the distance bias as the diameter of nodes. It means that two nodes cannot be located too close to avoid the infinite repulsive relation value between them. 
We define that node has attractive relations between its neighbours. By the Eq.~\ref{con:f_a}, the update of the attractive relation of node $k$ between neighbours is expressed as:
\begin{equation}
\nabla f_a(k) = -p * \sum_{i=1}^{Nei_k} \hat{w}_{ki}*\mathbf{d}_{ki},i \in Nei_k, 
\end{equation}
where $\hat{w}_{ki} \in \hat{\mathbf{W}}$ is the weight between node $k$ and node $i \in Nei_k$. $Nei_k$ is the neighbors set of $k$. $\mathbf{d}_{ki}$ is one element of matrix $\mathbf{D}_k =\{\mathbf{d}_{k1},\mathbf{d}_{k2},\ldots,\mathbf{d}_{k|V|-1}\}$. $\mathbf{D}_k \in \mathbf{R}^{\{|V|-1\}\times n}$ is defined as:
\begin{equation}
\mathbf{D}_k = {
\left[ \begin{array}{ccc}
\mathbf{u}_k - \mathbf{u}_1\\
\ldots \\
\mathbf{u}_k - \mathbf{u}_{k-1} \\
\mathbf{u}_k - \mathbf{u}_{k+1} \\
\ldots  \\
\mathbf{u}_k -\mathbf{u}_{|V|}
\end{array}
\right ]}.
\label{con:D_k}
\end{equation}
The $\nabla f_a(k)$ can be reformed as:
\begin{equation}
\nabla f_a(k)= -p * \sum_{i=1}^{Nei_k} \hat{w}_{ki}*(\mathbf{u}_k-\mathbf{u}_i).
\end{equation}
By the Eq.~\ref{con:f_r}, the update of repulsive relation between $k$ and other nodes is expressed as:
\begin{equation}
\nabla f_r(k) = q* \sum_{i = 1}^{|V|-1}\tilde{w}_{ki}* \frac{\mathbf{d}_{ki}}{\hat{d}_{ki}}, 
\end{equation}
$\hat{d}_{ki}$ is the $ith$ element of $\hat{\mathbf{D}}_k$, which is formulated as:
\begin{equation}
\hat{\mathbf{D}}_k = \mathbf{S}\cdot(\tilde{\mathbf{D}}_k^\top \circ \tilde{\mathbf{D}}_k^\top),
\label{con:hat_D_k}
\end{equation}
where $\circ$ is the Hadamard product of matrix. The matrix $\mathbf{S} = \left[1_1,1_2,\ldots,1_{n-1},1_n\right]$, which is used to sum columns of $\tilde{\mathbf{D}}_k^\top \circ \tilde{\mathbf{D}}_k^\top$. The notation $\tilde{\mathbf{D}}_k$ is the bias distance matrix, whose element is represented as:
\begin{equation}
\tilde{\mathbf{d}}_{ki} =|\mathbf{d}_{ki}|+\mathbf{b},
\end{equation}
where $\mathbf{d}_{ki}$ is an element of $\mathbf{D}_{k}$. So, $\nabla f_r(k)$ can be reformed as:
\begin{equation}
\nabla f_r(k) = q* \sum_{i = 1}^{|V|-1}\tilde{w}_{ki}* \frac{\mathbf{u}_{k}-\mathbf{u}_{i}}{||\mathbf{u}_k-\mathbf{u}_i|+\mathbf{b}|^2}.
\end{equation}

GForce directly updates the vectors through the nodes. In other words, each node has its own feature vector and updates itself, so the algorithm can learn and operate in parallel. We can assign the designated nodes to different CPUs, and each CPU calculates features of nodes independently. Nodes will reach a position where the forces are balanced after independent calculations according to nodes' relations. To measure whether all nodes are in equilibrium, we define the energy $E$ of the graph:
\begin{equation}
E = \sum_{k=1}^{|V|} ||\nabla f_a(k)+ \nabla f_r(k)||_2^2.
\label{con:energy}
\end{equation}
Symbol $E$ can be used to stop the iteration of learning progress. When $E$ is stable, it means that the node has reached its equilibrium position. In the proposed framework, each node retains its own feature vector $\mathbf{u}_k$, and updates it through the influence of relations imitating force in each iteration. The framework of GForce is shown in Fig~\ref{NetForce_framework6}. The pseudo-code of GForce is presented in Algorithm~\ref{NetForce_code}.

\begin{algorithm}[t]
\caption{GForce Algorithm}
\label{NetForce_code}
\begin{algorithmic}
\STATE {Input: graph $G = \{V,E\}$, randomly initial feature vectors $\mathbf{U}
\in \mathbf{R}^{|V|\times n}$ of nodes; }
\STATE {Output: feature vectors $\mathbf{U}$;}
\REPEAT
\FOR {$k \in |V|$}
\STATE{Calculate $\mathbf{D}_k$ by Eq.~\ref{con:D_k} and $\hat{\mathbf{D}}_{k}$ by
Eq.~\ref{con:hat_D_k}}
\STATE{$\nabla f_{r}(k) = q* \sum_{i = 1}^{|V|-1}
\frac{\mathbf{d}_{ki}}{\hat{\mathbf{d}}_{ki}} * \tilde{w}_{ki}$};
\STATE{$ \nabla f_{a}(k) = -p * \sum_{i=1}^{N_k} \mathbf{d}_{ki}*\hat{w}_{ki}$};
\STATE{$\nabla f(k) = \nabla f_{a}(k)+\nabla f_{r}(k)$};
\STATE{Calculate learning speed $h$};
\STATE{$\mathbf{u}_{k} = \mathbf{u}_{k} + h*\nabla f(k)$};
\ENDFOR
\STATE Move nodes in the feature space;
\UNTIL{Reached minimal energy state;}
\end{algorithmic}
\end{algorithm}

\section{Experiments}\label{sec:experiments}
In this section, we present a series of experimental analyses on GForce. We thoroughly evaluate the performance of our model by conducting extensive experiments analysis from three aspects including embedding analysis, label prediction, and graph visualization.
\begin{table}[t]
  \centering
  \caption{Datasets used in our experiments}
    \begin{tabular}{c|c|c|c|c|c}
\cline{1-6}    Dataset & Cora & Citeseer & Wikipedia & WebKB & 20newsgroups \\
    \hline
    |V| & 2,708 & 3,312 & 2,405 & 877 & 1,727 \\
    |E| & 5,429 & 4,732 & 17,981 & 1,605 & 2,982,529 \\
    |Y| & 7   & 6   & 20  & 5   & 20 \\
    Vertex attr & Yes & Yes & Yes & No  & No \\
    \hline
    \end{tabular}%
  \label{tab:datasets}%
\end{table}%
\begin{table}[t]
  \centering
  \caption{Summary of baselines}
    \begin{tabular}{l|c|c|c}
    \hline
    \multicolumn{1}{c|}{Methods} & First-order & Second-order & Higher-order \\
    \hline
    DeepWalk &  &  \checkmark   & \checkmark \\
    LINE & \checkmark & \checkmark &  \\
    Node2vec &     & \checkmark & \checkmark \\
    GraRep &     & \checkmark & \checkmark \\
    HOPE &     & \checkmark & \checkmark \\
    SDNE &  \checkmark & \checkmark  &   \\  
    \hline
    \end{tabular}%
  \label{tab:baselines}%
\end{table}%

\subsection{Experimental Setup}
GForce could run in parallel. We implement a multi-processes version of GForce, for which, each process could handle operations of a certain number of nodes. In the following experiments, we set the parameters $p = 1$ and $q = 5$. We use energy $E$ to determine whether the learning process is complete. If $E$ is stable, which means node has reached its equilibrium position and learning process is finished.

Before conducing experiments with real-world datasets, we first simply verify whether GForce follows the embedding results, that is, the two connected nodes in the feature space will be relatively close to each other, otherwise, nodes maintain far from each other. We learn feature vectors from a $15\times15$ grid graph, and then map the dimensionality reduction vector to nodes in 2D space to observe whether nodes follow the embedding results. Then we conduct an experiment on the dataset to verify the relationship between the embedding dimension and structure embedding accuracy.

To test the feature learning performance of GForce, we compare it with baselines on benchmark network datasets. These baselines are all graph structure-based embedding algorithm focusing on first-order proximity and second-order proximity. In the label prediction experiment, we train our algorithm on the citation network, and webpage network. We randomly sample a portion of the labeled nodes as training samples, and the rest as the test set. In each group of controlled experiments, we randomly select data from $40\%$ to $80\%$ of a set of labels and repeat 5 times. We compute the average Micro-F1 socre and Macro-F1 score for label prediction task. In the graph visualization experiment, we utilize t-SNE toolkit to reduce feature vectors extracted from the 20newsgroups dataset. Nodes of the same type are assigned the same color. Therefore, clustering the same nodes together can represent a better representation.
\subsubsection{Datasets}
Five different benchmark datasets including Cora, Citeseer, Wikipedia, WebKB, and 20newsgroups are selected to implement experiments. The overview of datasets is presented in Table~\ref{tab:datasets}.
\begin{itemize}
  \item Cora dataset is a citation network, which is one of the most commonly used benchmark dataset to verify the performance of network feature learning. In the dataset, each vertex (publication) is represented by a unique number. Links between vertexes (publications) are stored in the form of adjacency matrix or edge lists. Labels and attributes of vertexes are stored in two different dictionary files.
  \item Similar to Cora, Citeseer is a citation network that includes publications, publication classes, and links. Each attribute of the publication is extracted from the abstract. Attribute is represented by a 0/1-valued sequence. Each single value of attribute indicates whether the keyword exists in the abstract.
  \item Wikipedia dataset is a co-occurrence network extracted from Wikipedia website, containing keywords, their links and labels. This dataset has been pre-processed for ease of use. All nodes and labels are replaced with data in digital string format.
  \item WebKB (World Wide Knowledge Base) is a dataset of webpages. WebKB contains original hyperlinks of web pages. The data in the dataset are organized in a similar way to other datasets.
  \item 20newsgroups dataset is is one of the international standard data sets for text classification and text extraction. The dataset has about 20,000 newsgroup documents, which were evenly divided into 20 newsgroups with different topics.
\end{itemize}
\begin{figure}[t]
  \centering
  \includegraphics[width=3.5 in,height=3.7 in]{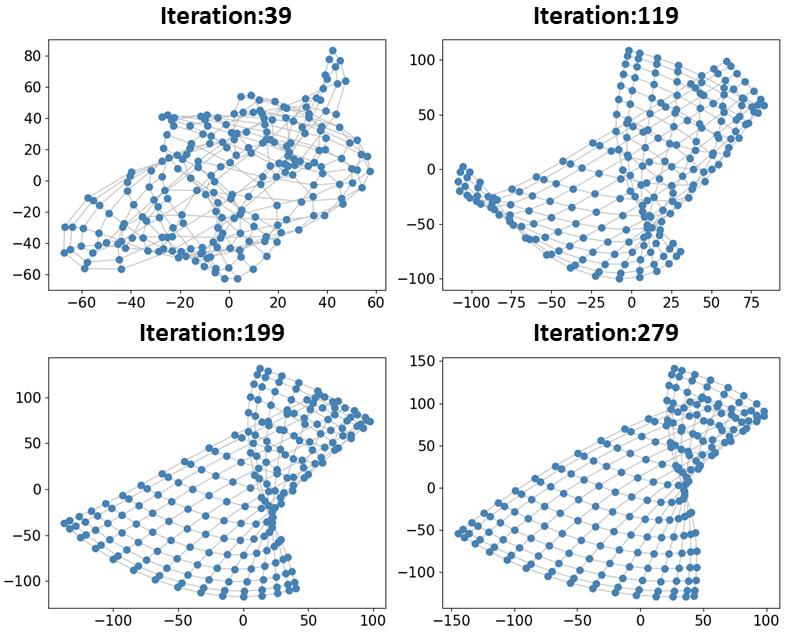}
  \caption{Verification embedding results of GForce on a grid graph.}
  \label{learning_process}
\end{figure}

\begin{figure}[t]
  \centering
  \includegraphics[width=3.0 in,height=4.5 in]{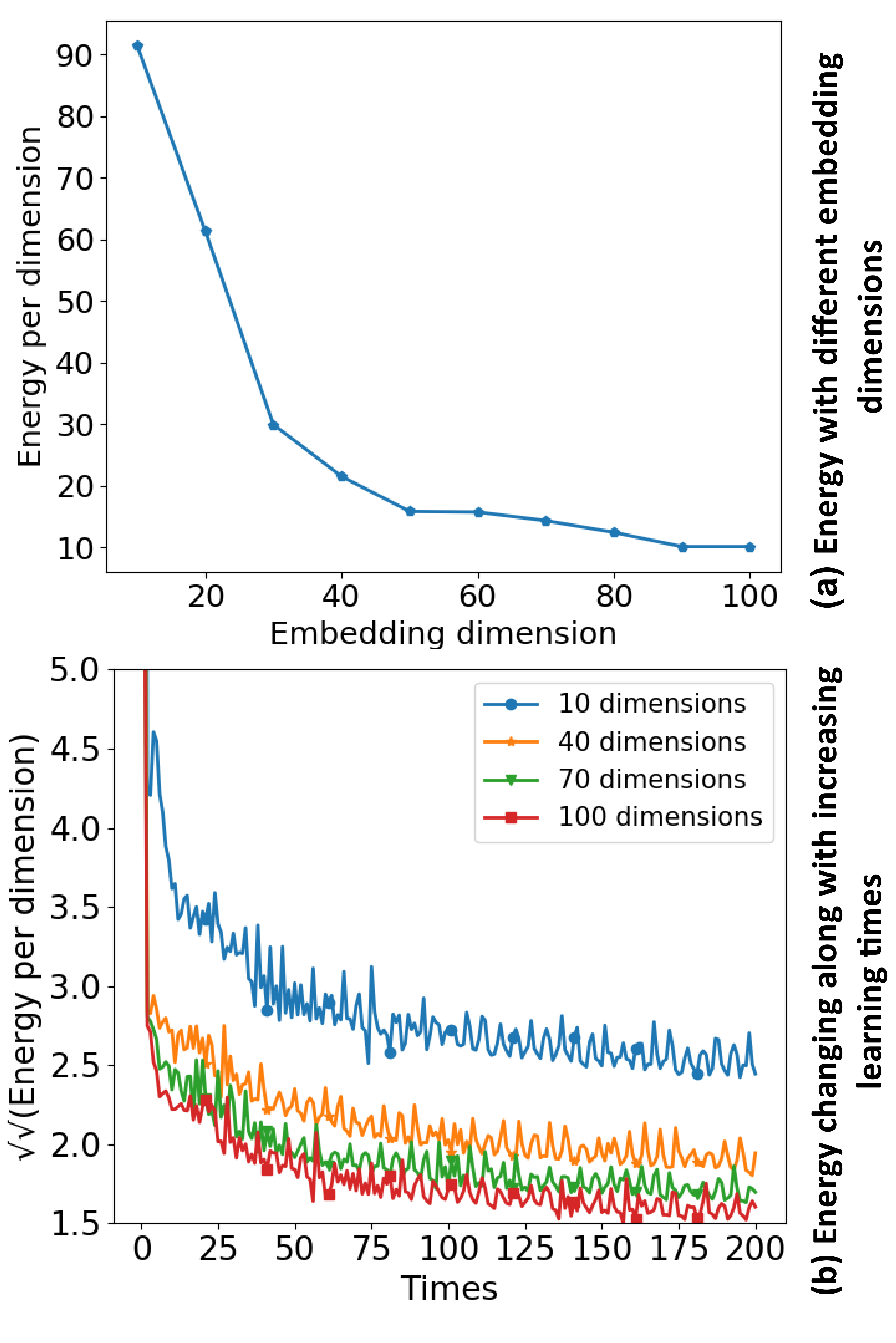}
  \caption{Relations between the dimension of feature space and energy. As shown in the top figure, high embedding dimension can achieve low energy under the same conditions. As shown in the bottom figure, the red line of 100 dimensions is consistently lower than that of other dimensions with the increasing learning time. Low energy means the corresponding dimensions of embedding space could effectively embed original graph structure information.}
  \label{energy_dim}
\end{figure}

\begin{table*}[htbp]
  \centering
  \caption{Results of label prediction on three benchmark datasets}
    \resizebox{\textwidth}{!}{
    \begin{tabular}{c|l|r|l|l|l|l|l|l|l|l|l|l|l|l|l|l}
    \hline
    \multicolumn{2}{c|}{\textbf{Datasets}} & \multicolumn{5}{c|}{\textbf{Cora}} & \multicolumn{5}{c|}{\textbf{ Citeseer}} & \multicolumn{5}{c}{\textbf{Wikipedia}} \\
    \hline
    \multicolumn{1}{l|}{\textbf{Maetric}} & \textbf{Algorithm} & \textbf{40\%} & \multicolumn{1}{r|}{\textbf{50\%}} & \multicolumn{1}{r|}{\textbf{60\%}} & \multicolumn{1}{r|}{\textbf{70\%}} & \multicolumn{1}{r|}{\textbf{80\%}} & \multicolumn{1}{r|}{\textbf{40\%}} & \multicolumn{1}{r|}{\textbf{50\%}} & \multicolumn{1}{r|}{\textbf{60\%}} & \multicolumn{1}{r|}{\textbf{70\%}} & \multicolumn{1}{r|}{\textbf{80\%}} & \multicolumn{1}{r|}{\textbf{40\%}} & \multicolumn{1}{r|}{\textbf{50\%}} & \multicolumn{1}{r|}{\textbf{60\%}} & \multicolumn{1}{r}{\textbf{70\%}} & \multicolumn{1}{r}{\textbf{80\%}} \\
    \hline
    \multirow{6}[2]{*}{\textbf{Micro-F1}} & DeepWalk & 0.761 & 0.770 & 0.780 & \multicolumn{1}{r|}{0.787} & \multicolumn{1}{r|}{0.789} & 0.576 & 0.578 & 0.589 & 0.590 & 0.594 & \textbf{0.672} & 0.680 & 0.683 & 0.687 & 0.688 \\
        & Node2vec & \multicolumn{1}{l|}{0.810} & 0.811 & 0.815 & 0.822 & 0.825 & 0.585 & 0.580 & 0.594 & 0.597 & 0.609 & 0.659 & 0.667 & 0.682 & 0.684 & 0.690 \\
        & LINE & \multicolumn{1}{l|}{0.755} & 0.773 & 0.778 & 0.797 & 0.797 & 0.493 & 0.500 & 0.501 & 0.507 & 0.527 & 0.645 & 0.666 & 0.652 & 0.650 & 0.652 \\
        & GraRep & \multicolumn{1}{l|}{0.768} & 0.769 & 0.781 & 0.803 & 0.785 & \multicolumn{1}{r|}{0.548} & 0.545 & 0.542 & 0.545 & 0.545 & 0.647 & 0.651 & 0.656 & 0.662 & 0.659 \\
        & HOPE & \multicolumn{1}{l|}{0.644} & 0.652 & 0.654 & 0.676 & 0.656 & \multicolumn{1}{r|}{0.452} & 0.448 & 0.453 & 0.457 & 0.452 &  0.591   & 0.606    &  0.611   &  0.599   & 0.609 \\
        & \textbf{GForce} & \multicolumn{1}{l|}{\textbf{0.817}} & \textbf{0.832} & \textbf{0.837} & \textbf{0.838} & \textbf{0.865} & \textbf{0.643} & \textbf{0.648} & \textbf{0.649} & \textbf{0.648} & \textbf{0.647} & 0.626 & \textbf{0.685} & \textbf{0.689} & \textbf{0.699} & \textbf{0.700} \\
    \hline
    \multirow{6}[2]{*}{\textbf{Macro-F1}} & DeepWalk & \multicolumn{1}{l|}{0.732} & \multicolumn{1}{r|}{0.765} & \multicolumn{1}{r|}{0.768} & \multicolumn{1}{r|}{0.776} & \multicolumn{1}{r|}{0.782} & 0.525 & 0.528 & 0.541 & 0.528 & 0.544 & 0.568 & \textbf{0.580} & 0.592 & 0.585 & 0.590 \\
        & Node2vec & \multicolumn{1}{l|}{0.776} & 0.806 & 0.808 & 0.812 & 0.817 & 0.545 & 0.531 & 0.543 & 0.554 & 0.572 & \textbf{0.649} & 0.547 & 0.548 & 0.561 & 0.559 \\
        & LINE & \multicolumn{1}{l|}{0.715} & 0.762 & 0.766 & 0.794 & 0.801 & 0.463 & 0.462 & 0.461 & 0.470 & 0.483 & 0.572 & 0.548 & 0.558 & 0.570 & 0.568 \\
        & GraRep & \multicolumn{1}{l|}{0.746} & 0.745 & 0.761 & 0.786 & \multicolumn{1}{r|}{0.777} & 0.485 & 0.480 & 0.574 & 0.475 & 0.482 & 0.498 & 0.507 & 0.507 & 0.522 & 0.519 \\
        & HOPE & \multicolumn{1}{l|}{0.626} & 0.630 & 0.632 & 0.652 & 0.634 & 0.392 & 0.386 & 0.393 & 0.392 & 0.394 &  0.426   &  0.438   &  0.440   & 0.433    &  0.432\\
        & \textbf{GForce} & \textbf{0.778} & \textbf{0.818} & \textbf{0.821} & \textbf{0.827} & \textbf{0.854} & \textbf{0.595} & \textbf{0.595} & \textbf{0.590} & \textbf{0.595} & \textbf{0.594} & 0.457 & 0.570 & \textbf{0.602} & \textbf{0.615} & \textbf{0.618} \\
    \hline
    \end{tabular}%
    }
  \label{tab:classfication_results}%
\end{table*}%

\subsubsection{Baseline Methods}
We validate the performance of GForce by comparing it with graph structure-based baselines including DeepWalk~\cite{perozzi2014deepwalk}, LINE~\cite{tang2015line}, Node2vec~\cite{grover2016node2vec}, GraRep~\cite{cao2015grarep}, HOPE~\cite{ou2016asymmetric} and SDNE~\cite{wang2016structural}. DeepWalk, Node2vec, GraRep, and HOPE focus on global graph characteristics including second-order and higher-order proximities. LINE can simultaneously maintain the first-order and second-order proximity of the graph. The properties of them are summarized in Table~\ref{tab:baselines}. The detailed descriptions of these methods are listed as follows.
\begin{itemize}
\item DeepWalk~\cite{perozzi2014deepwalk} algorithm is established on word2vec model. The algorithm utilizes random walk to sample graph and then generates nodes sequence, that is similar to word sequence. DeepWalk could keep the local and global graph characteristics.
\item LINE~\cite{tang2015line} is a kind of network embedding learning algorithm for large graphs, which can preserve the first-order and second-order proximity of the graph.
\item Node2vec~\cite{grover2016node2vec} is an improved algorithm based on DeepWalk. The algorithm is combined with biased sampling strategy with parameters $p$ and $q$ so as to keep local or global characteristics of the graph.
\item GraRep~\cite{cao2015grarep} also leverages the skip-gram model~\cite{mikolov2013efficient} but keeps the high-order proximity of graph. The main idea is that nodes with k-steps neighbors should have similar representations in the embedding space.
\item HOPE~\cite{ou2016asymmetric} could capture the asymmetric high-order proximity in a directed graph, which is a kind of matrix factorization based network embedding algorithm.
\item SDNE~\cite{wang2016structural} could preserve first-order and seconder-order proximity of nodes. It is a network representation learning model based on the deep autoencoder approach.
\end{itemize}

\subsection{Quantitative Results}
\begin{figure*}[t]
  \centering
  \includegraphics[width=7.2 in,height=1.6 in]{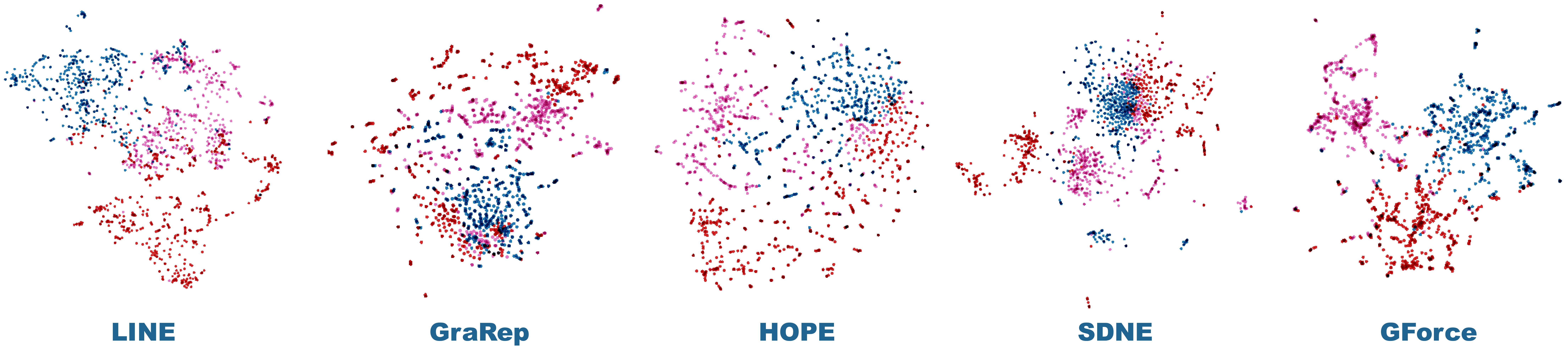}
  \caption{Visualization results of 20newsgroups dataset. Positions of vertexes are two-dimensional vectors reduced by t-SNE from the learned feature vectors. Color of a vertex indicates the category of the webpage.}
  \label{visualization}
\end{figure*}

The experimental results are summarized from Fig~\ref{learning_process} to Fig~\ref{visualization} and Table~\ref{tab:classfication_results}. We can make the following observations from the results.
\subsubsection{Embedding Analysis}
To verify that GForce is able to effectively embed structure information, we first conduct embedding analysis experiments. In the following, we present verification purposes and corresponding results.

\textbf{Structure Embedding Verification}. GForce follows the principle of spring-electrical model. To ensure the reliability of the structure embedding, we utilize GForce to extract structure information of $15 \times 15$ grid graph. To facilitate the observation of embedded results, we map nodes with vectors to 2D feature space with an increasing frame.

From the Fig~\ref{learning_process}, we can see that nodes are in chaos when the iteration time is small. As the number of learning iterations increases, unconnected nodes gradually move away from each other. Therefore, the nodes gradually learn the correct vector to become a grid graph. Experimental results show that GForce follows the aforementioned embedding hypothesis and can be used for structural feature extraction.

\textbf{Embedding Accuracy with Dimensions}. The feature learning of GForce maps nodes into a relatively high continuous feature space $(n\gg3)$, because high dimensional feature space could effectively represent features. To prove this conclusion, we verified it through the WebKB dataset. The results are shown in Fig~\ref{energy_dim}.

From the Fig~\ref{energy_dim}(a), the energy per dimension of node decreases with the increasing dimension of feature space, which means that the energy of nodes in high-dimensional feature space is lower. The low energy indicates that nodes have reached the equilibrium position in the feature space. From Fig~\ref{energy_dim}(b), the energy of dimension in 100-dimensions feature space is consistently lower than others with increasing learning times. In conclusion, high-dimensional feature space can better preserve the original structure information.

\subsubsection{Label Prediction}
We conduct label prediction on three benchmark datasets. The results are shown in Table~\ref{tab:classfication_results}.

From the Table~\ref{tab:classfication_results}, we can see a significant improvement in GForce compared with baselines on the Cora dataset with both Micro-F1 and Macro-F1 score. As the training ratio increases, GForce obtains higher accuracy. Node2vec is based on DeepWalk and obtains the best performance among baseline algorithms. The prediction accuracy of GForce is consistently better than Node2vec on both Micro-F1 and Macro-F1. Compared with Node2vec, the best performance of GForce in Micro-F1 and Macro-F1 exceed $0.040$ and $0.037$ respectively. DeepWalk and LINE produce similar results, and can always improve prediction accuracy on HOPE.

From the Table~\ref{tab:classfication_results} of label prediction results on Citeseer dataset, we can observe that the Micro-F1 score of GForce is consistently exceeding $0.6$, but other baselines are not. In this experiment, the performance of baselines is similar to previous results. Node2vec performs best among the baselines. In addition, DeepWalk is slightly better than GraRep in both Micro-F1 and Macro-F1 scores.

From the Table~\ref{tab:classfication_results} of label prediction results on Wikipedia dataset, DeepWalk achieved $0.672$ on Micro-F1 score at 40\%, which is over other baselines and GForce. However, node2vec achieved $0.649$ on Macro-F1 score at 40\%, that is better than other baselines and GForce. GForce obtained higher accuracy in Micro-F1 at 50\% than others, and exceed baselines on both Micro-F1 and Macro-F1 with higher proportions of training set.  

\subsubsection{Graph Visualization}
We select $3$ classes of documents from the 20newsgroups dataset. Different colors in figures indicates different categories.

As can be seen from Fig~\ref{visualization}, the layouts using SDNE and GraRep are not very effective because most different types of nodes are mixed. Visualization results of HOPE and LINE are better because nodes are clustered in separable groups with clear boundaries. However, most nodes of the same type are not close in the same group. For GForce, different types of nodes are relatively concentrated in distinguishable regions, and each group has a clear boundary. Therefore, from the visualization results, GForce achieves better performance than other baselines.
\section{Conclusion}\label{sec:conclusion}
In this paper, we propose GForce algorithm, which can better represent the relations among nodes in network. The main two steps of GForce are attractive relation step and repulsive relation step, which can be a (non)linear method to capture hidden relationship between nodes. GForce is different from most network representation learning models that rely on global optimal solutions. It can learn the feature vector by the node itself. This feature makes the algorithm own the ability of parallel computing. The main advantage of GForce is that it can fully capture the feature of the graph. Empirically, we verify the performance of GForce through the tasks of label prediction and graph visualization on real-world network datasets. The results show that our algorithm achieves better performance than state-of-the-art baselines. In the future, gravity method will be taken into consideration in our proposed model, which could map the node towards a high degree node (a fix point) in the feature space. In order to further reduce computational complexity, edge-cutting strategy will also be considered. Learning each subgraph individually can effectively reduce learning complexity and shorten learning time.


\normalem
\bibliographystyle{IEEEtran}
\bibliography{IEEEabrv,references}

\end{document}